\documentclass[twocolumn]{article} % Use twocolumn document class
\usepackage{arxiv}
\usepackage[square,sort,comma,numbers]{natbib}
\usepackage[utf8]{inputenc} % allow utf-8 input
\usepackage[T1]{fontenc}    % use 8-bit T1 fonts
\usepackage[shortlabels]{enumitem}
\usepackage[obeyspaces]{url}
\usepackage{wrapfig}
\usepackage{booktabs}       % professional-quality tables
\usepackage{amsfonts}       % blackboard math symbols
\usepackage{nicefrac}       % compact symbols for 1/2, etc.
\usepackage{microtype}      % microtypography
\usepackage{graphicx}
\usepackage{natbib}
\usepackage{doi}
\usepackage{multicol}
\usepackage{multirow}
\usepackage{tabularx}
\usepackage{tabulary}
\usepackage{makecell}
\usepackage{tabularx,ragged2e}
\usepackage{fancyhdr, blindtext}
\usepackage{subfigure}
\usepackage{float}
\usepackage{hyperref}
\usepackage{hhline}
\usepackage{array}
\usepackage{amsmath}
\usepackage{diagbox}
\usepackage{color}
\usepackage{textcomp}
\usepackage{comment}
\usepackage{relsize}
\usepackage[symbol]{footmisc}
\usepackage{caption}
\usepackage{chngcntr}
\usepackage{amssymb}%
\usepackage{amsthm}%
\usepackage{mathrsfs}%
\usepackage[title]{appendix}%
\usepackage{xcolor}%
\usepackage{textcomp}%
\usepackage{manyfoot}%
\usepackage{booktabs}%
\usepackage{algorithm}%
\usepackage{algorithmicx}%
\usepackage{algpseudocode}%
\usepackage{listings}%
\usepackage{natbib} 
\usepackage{comment}
\usepackage{etex} % Extend TeX memory
\usepackage{morewrites}
\usepackage{authblk} % For managing author affiliations
\usepackage[capitalise]{cleveref}

\renewcommand{\headeright}{Generative AI in Surgery}
\renewcommand{\shorttitle}{CoSimGen: Controlled Image-Mask Generation}
\renewcommand{\undertitle}{Advances in Generative AI}

\setcitestyle{square}
\hypersetup{
pdftitle={CoSimGen},
pdfsubject={q-bio.NC, q-bio.QM},
pdfauthor={Bose et al.},
pdfkeywords={Diffusion models, surgical image-label synthesis, paired image-mask generation, segmentation, controlled generation},
}

\title{CoSimGen: Controllable Diffusion Model for Simultaneous Image and Mask Generation}

\date{}

\author{ \normalsize
    \textbf{Rupak Bose}\textsuperscript{1,†}, 
    \textbf{Chinedu Innocent Nwoye}\textsuperscript{1,2,†,*},
    \textbf{Aditya Bhat}\textsuperscript{1,†}, 
    \textbf{Nicolas Padoy}\textsuperscript{1,2}\\
    \textsuperscript{1}ICube, UMR7357, CNRS, INSERM, University of Strasbourg, France\\
    \textsuperscript{2}IHU Strasbourg, France\\
    {\smaller
        \textsuperscript{†}{Equal contribution (co-first author)}. \quad      \textsuperscript{*}Corresponding author: {\tt nwoye@unistra.fr}\\        
        {\normalsize Project page:~~\tt \url{https://camma-public.github.io/endogen/cosimgen}}        
    }
}

\begin{document}

\twocolumn[{%
\renewcommand\twocolumn[1][]{#1}%
\maketitle
\vspace*{-10mm}

%%%%%%%%%%%%%%%% ABSTRACT %%%%%%%%%%%%%%%%%%%%
\begin{abstract}  
The acquisition of annotated datasets with paired images and segmentation masks is a critical challenge in domains such as medical imaging, remote sensing, and computer vision. Manual annotation demands significant resources, faces ethical constraints, and depends heavily on domain expertise. Existing generative models often target single-modality outputs, either images or segmentation masks, failing to address the need for high-quality, simultaneous image-mask generation. Additionally, these models frequently lack adaptable conditioning mechanisms, restricting control over the generated outputs and limiting their applicability for dataset augmentation and rare scenario simulation.  
We propose \textit{CoSimGen}, a diffusion-based framework for controllable simultaneous image and mask generation. Conditioning is intuitively achieved through (1) text prompts grounded in class semantics, (2) spatial embedding of context prompts to provide spatial coherence, and (3) spectral embedding of timestep information to model noise levels during diffusion. To enhance controllability and training efficiency, the framework incorporates contrastive triplet loss between text and class embeddings, alongside diffusion and adversarial losses. Initial low-resolution outputs (\(128 \times 128\)) are super-resolved to \(512 \times 512\), producing high-fidelity images and masks with strict adherence to conditions.  
We evaluate CoSimGen on metrics such as FID, KID, LPIPS, Class FID, Positive predicted value for image fidelity and semantic alignment of generated samples over 4 diverse datasets.
CoSimGen achieves state-of-the-art performance across all datasets, achieving the lowest KID of 0.11 and LPIPS of 0.53 across datasets.
% By addressing key gaps in simultaneous generation and precision-driven applications, CoSimGen advances cross-domain AI research and reduces dependence on manual annotation.  
\end{abstract}

\vspace*{2mm}
\keywords{: Generative AI \and  diffusion model\and  segmentation \and  dataset\and  image-mask generation \and inception distance metrics\\[0.05in]}

\vspace*{5mm}
}]

% \vspace{-2em}
\enlargethispage{\baselineskip}

%%%%%%%%%%%%%%%%%%%% SECTIONS %%%%%%%%%%%%%%%%%%%%%%%%

\section{Introduction}
\label{sec:intro}

\begin{figure}
    \centering
    \includegraphics[width=0.9\linewidth]{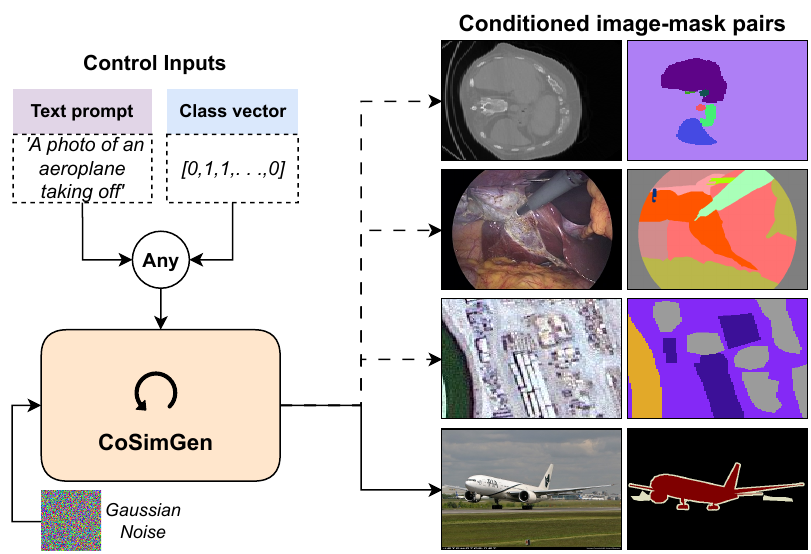}
    \caption{CoSimGen takes either text or class vector as input prompt and generates a high-resolution image minimally representing the prompt context and a mask segmenting all the objects in the prompt.}
    \label{fig:teaser}
\end{figure}

The generation of paired image and segmentation mask data is a crucial task for various applications, ranging from training machine learning models to assisting human learning through demonstrations, simulations, and interactive systems. In domains such as medical imaging~\cite{aggarwal2011role}, autonomous driving~\cite{feng2020deep}, surgical intervention~\cite{nwoye2021deep}, geospatial analysis~\cite{akccay2008automatic}, security and surveillance~\cite{ojha2015image}, generating high-quality annotated datasets can be prohibitively expensive and labor-intensive. Recent advances in generative models, including Variational Autoencoders (VAEs) \cite{kingma2013auto}, Generative Adversarial Networks (GANs) \cite{goodfellow2014generative}, and Diffusion Models \cite{sohl2015deep}, have made significant strides in image generation. However, most methods focus on individual modalities, either images~\cite{ramesh2021zero,isola2017image} or segmentation masks~\cite{text4seg2024,seggenerator2023}, without addressing the need for high-quality, simultaneous generation of both image and mask pairs~\cite{bhat2025simgen}. Furthermore, these models~\cite{wu2023diffumask} often lack flexibility in conditioning on multiple input signals, limiting control over the generated outputs.

In this work, we propose \textit{CoSimGen}, a diffusion-based unified framework for \textbf{Co}ntrollable \textbf{S}imultaneous \textbf{I}mage and segmentation \textbf{M}ask \textbf{Gen}eration, conditioned on either textual prompts or class labels to enable fine-grained control during training and inference (\cref{fig:teaser}). CoSimGen leverages a Denoising Diffusion Probabilistic Model (DDPM)~\cite{ho2020denoising} as its base technology, utilizing a Conditional U-Net~\cite{ronneberger2015u} architecture with skip connections to preserve high-resolution features. 
We introduce a novel fusion methodology, {\textit{Spectron}} (spatio-spectral embedding fusion), which integrates conditions with U-Net features in both spectral and spatial domains. Specifically, we fuse class features along the spatial axis to capture shape and texture, while timestep features are fused along the channel (spectral) axis, as they estimate the input noise level, assumed uniform across channels.
In addition, we introduce {\textit{Textron}} (text-grounded semantic conditioning), aligning class features with text features, enabling \textit{`hot-swapping'} for text-based generation of image-mask pairs during inference.
During training, the framework employs contrastive learning to approximate text and class embeddings, similar to approaches like CLIP \cite{radford2021learning}, enabling seamless conditioning on either modality during inference - a unique contribution in this work. 
In addition to the diffusion loss and contrastive loss, we employ an adversarial loss that acts as a regularizer.
Initial outputs at \(128 \times 128\) resolution are super-resolved to \(512 \times 512\), ensuring high-fidelity images and accurate masks. By conditioning on user-specified inputs, CoSimGen achieves strict adherence of masks to conditions while maintaining variability in image generation, enhancing robustness and adaptability across diverse datasets.

We evaluate {CoSimGen} on four diverse segmentation datasets, including Cholecseg8k \cite{hong2020cholecseg8k}, BTCV~\cite{gibson_2018_1169361}, MBRSC Semantic Segmentation~\cite{humansintheloop2024}, and PASCAL-VOC \cite{pascal-voc-2012}.
These datasets cover a wide array of domains, such as surgical training, medical imaging, general computer vision, and geospatial, demonstrating the adaptability and robustness of our approach across domains. 
To assess the quality of the generated images, we use Fréchet Inception Distance (FID) \cite{yu2021frechet}, Kernel Inception Distance (KID) \cite{binkowski2018demystifying}, VGG distance and Learned Perceptual Image Patch Similarity (LPIPS) as our primary evaluation metrics. 
To evaluate the alignment of generated segmentation masks with their corresponding images, we use \textit{Semantic Fréchet Inception Distance (sFID)}~\cite{bhat2025simgen} and Positive Predictive Value (PPV) as our primary metrics.

Our extensive experimental results demonstrate that {CoSimGen} outperforms existing baselines on all datasets on most metrics. Our approach significantly improves the fidelity of generated paired image-mask data, offering new possibilities for creating realistic simulation environments for both machine learning and human-oriented training tasks. Through our method, we provide a scalable and flexible solution to data generation, offering fine control over the creation of domain-specific datasets with precise annotations. 
It is particularly useful for applications like surgical training, where specific regions of the image (e.g., organ tissues, surgical tools) must be generated with high accuracy.
Also, the capability of the model to generate unlimited data makes it a decent source for model pretraining and domain adaptation.

\section{Related Work}
\label{sec:literature}

\noindent\textbf{Image generation: from unconditional to conditional synthesis.~}  
Generative models have evolved significantly, starting with Variational Autoencoders (VAEs) \cite{kingma2013auto} and Generative Adversarial Networks (GANs) \cite{goodfellow2020generative}, which laid the foundation for image synthesis. GAN variants, such as Pix2Pix \cite{isola2017image} and StyleGAN \cite{karras2019style}, advanced conditional generation by leveraging paired data or domain-specific priors. Recently, Diffusion Models \cite{ho2020denoising, rombach2022high} have surpassed traditional methods, offering improved diversity and fidelity in image generation tasks, such as text-to-image synthesis (e.g. DALL·E \cite{ramesh2021zero}, Imagen~\cite{saharia2022photorealistic}, Surgical Imagen~\cite{nwoye2025surgical}). However, most of these methods are limited to generating either images or single modalities, often neglecting simultaneous outputs like segmentation masks.

\noindent\textbf{Segmentation and paired data generation.~}  
In segmentation, paired data are crucial for supervised learning, yet acquiring such data is labor-intensive and domain-constrained. Recent efforts like Text4Seg \cite{text4seg2024} and SegGen \cite{seggenerator2023} attempt to synthesize segmentation data indirectly. Text4Seg maps text to image segmentation datasets, while SegGen generates segmentation masks conditioned on textual descriptions. In medical imaging, methods like HVAE \cite{autoencoding2023} focus on autoencoding paired data but often lack generalizability or fine-grained control. These approaches either treat image and mask generation as separate tasks or rely heavily on predefined pipelines. This limits adaptability. 

\begin{figure*}[t]
    \centering
    \includegraphics[width=0.9\linewidth]{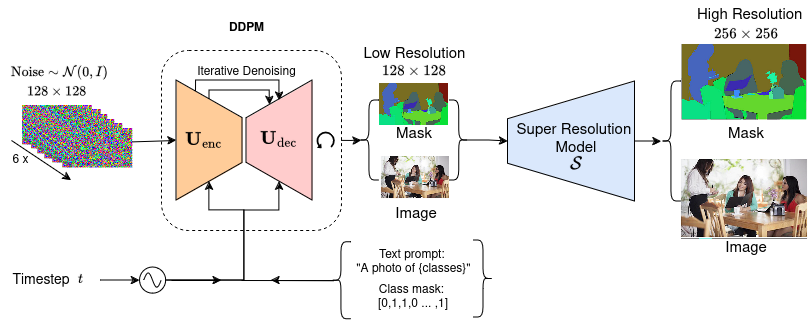}
    \caption{Architecture of CosimGen - showing the input conditioning, the diffusion process and super resolution of the generated outputs.}
    \label{fig:arch}
\end{figure*}

\noindent\textbf{Simultaneous generation of image and segmentation masks.~}
While prior works explore conditional generation in isolation, generating paired images and segmentation masks in a single process remains underexplored. CoModGAN \cite{zhao2021comodgan} attempts image-masked completion but is task-specific and does not generalize across domains. Similarly, diffusion-based models \cite{ho2020denoising, rombach2022high} have yet to address simultaneous generation across modalities. OVDiff~\cite{marcos2024open} generates images and rely on input vocabularies to segment requested objects. SimGen~\cite{bhat2025simgen} simultaneously generates image and masks with no input conditioning and is explored only on surgical datasets. DiffuMask~\cite{wu2023diffumask} and SatSynth~\cite{toker2024satsynth} generate paired image and pixel-level labels conditioned only on the textual modality and explore on a single domain. 
These gaps limit the development of models capable of efficiently generating high-fidelity, multimodal outputs for diverse applications, from healthcare to remote sensing and general computer vision.

\section{Methods}
\label{sec:methods}

% What is your task objective?
Our goal is to generate paired images and semantic segmentation masks, guided by user input prompts, for synthetic data creation and educational purposes.
To achieve this, we propose a Controllable Simultaneous Image-Mask Generator (CoSimGen), a diffusion-based framework that utilizes contrastive learning to seamlessly condition generation on text or class labels, ensuring precise alignment between images and masks in a unified process.

\subsection{Task Formalization}  
\label{sec:formalization}
Let \( \mathcal{D} = \{(\mathbf{X}_i, \mathbf{y}_i)\}_{i=1}^n \), where \( \mathbf{X}_i \in \mathbb{R}^{C \times H \times W} \) represents an image, and \( \mathbf{y}_i \in \{0, 1\}^{H \times W} \) is the corresponding segmentation mask. The mask \( \mathbf{y}_i \) contains the segmented objects, which are associated with class labels from \( \mathcal{C} = \{c_1, c_2, \dots, c_k\} \).  
The conditioning vector \( \mathbf{c}_i \) is derived from the mask \( \mathbf{y}_i \) and encodes the present classes. Alternatively, a text prompt \( \mathbf{t}_i \) can be provided as a caption of the image, limited to describing only the objects present in the segmentation mask.  
The task is to train a model \( \mathcal{M} \) that generates image-mask pairs \( (\hat{\mathbf{X}}, \hat{\mathbf{y}}) \) simultaneously, conditioned on the class labels \( \hat{\mathbf{c}} \) and/or the text prompt \( \hat{\mathbf{t}} \), such that the generated segmentation mask \( \hat{\mathbf{y}} \) aligns with the generated image \( \hat{\mathbf{X}} \) and both are similar to real samples from \( \mathcal{D} \). The goal is to maximize the likelihood of generating paired data that closely resembles real samples, conditioned on the class labels or the textual description of the segmented objects.

\subsection{Model Architecture}  
The proposed CoSimGen is built on the foundation of diffusion process to synthesize clean outputs by iteratively denoising the input features. The architecture, illustrated in \cref{fig:arch}, consists of:  
(a) a low-resolution (LR) generator that ensures spatial coherence between the image-mask pair and the queried class,  
(b) a spatio-spectral embedding fusion to seamlessly incorporates text/class embeddings and timestep embeddings into the model features,  
(c) a text-grounded semantic conditioning mechanism enabling flexible user-specified control, and  
(d) a super-resolution (SR) module to upscale LR outputs to high-resolution (HR) spatial dimensions.

\subsection{Low-Resolution Image-Mask Pair Generation}  
We employ a Conditional U-Net~\cite{ronneberger2015u} within a conditional Denoising Diffusion Probabilistic Model (DDPM)~\cite{ho2020denoising} to generate low-resolution image-mask pairs. 
The U-Net facilitates the generation of coherent image-mask pairs by conditioning on class embeddings derived from text priors. Specifically, the model takes the following inputs:
\begin{enumerate}  
    \item Noisy image-mask pair $\mathbf{X}_t$, where $t$ indicates the noise level in the diffusion process.  
    \item Binary mask condition $\mathbf{M} \in \{0, 1\}$, where $m = 1$ specifies the presence of the queried class.  
    \item Queried text prompt $\mathbf{Z}_q$, such as "A photo of \{classes\}".  
    \item Diffusion timestep $\mathbf{t}$, represents the noise level in $\mathbf{X}_t$.  
\end{enumerate}

\subsubsection{Text encoder}  
The text encoder $\mathcal{E}_z$ processes the queried text $\mathbf{Z}_q$ by passing it through a frozen sentence transformer, followed by a series of linear transformations that project the input text into a \( D \)-dimensional embedding space. 
This whole process can be expressed as $\mathbf{Z}_{\text{emb}} \in \mathbb{R}^{1 \times D}  = \mathcal{E}_z(\mathbf{Z}_q, \theta_T)$.

\subsubsection{Class encoder}
The class encoder $\mathcal{E}_c$ contains a learnable weight matrix $\mathbf{W}_c \in \mathbb{R}^{c \times d}$, where $c$ is the number of classes and $d$ is the feature dimension. The encoder takes a condition mask $\mathbf{M} \in \{0, 1\}^c$, where a value of $1$ at an index indicates the presence of that specific class.
To obtain the condition representation, the mask $\mathbf{M}$ is element-wise multiplied with $\mathbf{W}_c$ such that $\mathbf{C}_{\text{masked}} = \mathbf{M} \odot \mathbf{W}_c$. Then, a sum is taken along the class axis to aggregate features from the non-masked classes, resulting in a represented condition feature vector $\mathbf{C}_{\text{feat}} \in \mathbb{R}^{1 \times f}$ such that $\mathbf{C}_{\text{feat}} = \sum_{i=1}^{c} \mathbf{C}_{\text{masked}}[i, :]$. 
This $\mathbf{C}_{\text{feat}}$ passes through a series of linear transformations projecting it into a $D$-dimensional class embedding. We represent this as $\mathbf{C}_{\text{emb}} \in \mathbb{R}^{1 \times D} = \mathcal{E}_c(M, \theta_{\mathcal{E}_c})$.\\

\subsubsection{Timestep encoder}
The timestep encoder $\mathcal{E}_t$ takes in the time $t \in \mathbb{R}^{1 \times 1}$ at which the noisy input $ \mathbf{X}_t $ is sampled. This time value $t$ is first processed using a sinusoidal embedding function to project it into a feature dimension, resulting in an intermediate timestep embedding.
Then, this projected feature undergoes a sequence of linear transformations and is finally projected into a $D$-dimensional space to obtain the final timestep embedding. We represent this as $T_{\text{emb}} \in \mathbb{R}^{1 \times D} = \mathcal{E}_t(:, \theta_{\mathcal{E}_t})$.

\subsubsection{Conditional UNet}
The Conditional U-Net, denoted as $\mathbf{U}$, is based on a Residual U-Net~\cite{ronneberger2015u} architecture. It consists of an encoder, $\mathbf{U}_{\text{enc}}$, that performs sequential downsampling until it reaches a bottleneck feature representation, and a decoder, $\mathbf{U}_{\text{dec}}$, that sequentially upscales these feature maps back to the original spatial resolution. Residual connections are added between encoder and decoder feature maps at matching spatial resolutions to allow information flow between corresponding layers. 
To make the U-Net conditional, we inject the conditions $\mathbf{C}$ into the feature maps at all resolutions. To obtain text-aligned features for each resolution, the conditioning styles: (a) {spatio-spectral embedding fusion (Spectron)} and (b) {text-grounded semantic conditioning (Textron)} are employed. Following this, the conditioned features ensure that the U-Net is aligned with the textual context throughout the encoding-decoding process during training whereas during inference we can just switch to using text for generation of image mask pairs.

\subsection{Spatio-Spectral Embedding Fusion (Spectron)}
\begin{figure}
    \centering
    \includegraphics[width=1.23456789\linewidth]{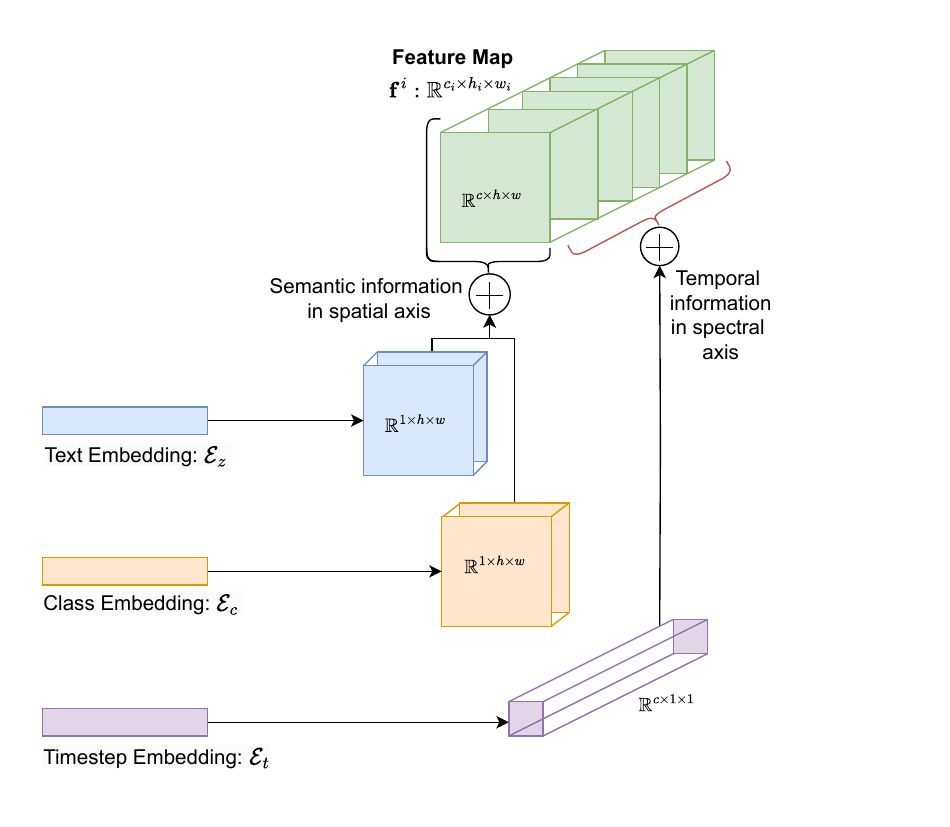}
    \caption{Spectron: spatio-spectral embedding fusion for semantic context and temporal feature conditioning.}
    \label{fig:spatiospectral}
\end{figure}
In traditional generative models, conditioning feedback is applied by direct concatenation along the latent space or by adding conditions to feature maps along the channel axis. Diffusion models often follow a similar approach, where the conditional embeddings and timestep embeddings are introduced by adding them to the channel axis of the feature maps. While effective, this approach does not fully exploit the semantic richness of the conditional embeddings. 

To bridge this gap, we introduce \textit{Spectron} \cref{fig:spatiospectral}, a strategy that injects conditions into feature maps at all resolutions, allowing for a more intuitive conditioning process. Recognizing that class conditions, such as the class embedding $\mathbf{C}_{\text{emb}}$, represent a semantic understanding of the image and mask, we propose spatially embedding this information. This semantic representation governs the shape, outline, and textures within the generated image and mask. Therefore, it is intuitively powerful to apply the semantic conditional vectors along the spatial dimensions, thereby spatially conditioning the features $\mathbf{f}$ at each resolution $i$:
\begin{equation} \label{eq:spatialmix}
\mathbf{f}_{\text{cond}}^{i, \text{spatial}} = \mathbf{f}^i + \mathbf{C}^i_{\text{emb}}
\end{equation}
where $ \mathbf{f}^i:\mathbb{R}^{c_i \times h_i \times w_i}$ and $ \mathbf{C}_{\text{emb}}^i:\mathbb{R}^{1 \times h_i \times w_i}$ thus adding the conditional embedding in the spatial dimension.

The timestep embedding $\mathbf{T}_{\text{emb}}$, by contrast, encodes the noisiness of the input, and the noise level is assumed to affect all channels uniformly and equally. Hence, it becomes intuitive to apply the timestep conditioning along the channel dimension of the noisy feature maps, thereby spectrally conditioning the features $\mathbf{f}$ at each resolution $i$: 
\begin{equation} \label{eq:spectralmix}
\mathbf{f}_{\text{cond}}^{i, \text{spectral}} = \mathbf{f}_{\text{cond}}^{i, \text{spatial}} + \mathbf{T}^i_{\text{emb}}
\end{equation}
where $ \mathbf{f}_{\text{enc}}^{i, spatial}:\mathbb{R}^{c_i \times h_i \times w_i}$ and $ \mathbf{T}_{\text{emb}}^i:\mathbb{R}^{c_i \times 1 \times 1}$.
% thus adding the conditional embedding in the spatial dimension.
By combining these two perspectives, Spatio-Spectral Feature Mixing enables both semantic and temporal feedback, allowing $\mathbf{U}$ to generate features that are spatially aligned with the condition semantics and spectrally aligned with the temporal noise level. This dual conditioning mechanism ensures that the U-Net captures a deep alignment between the class condition and timestep information across all spatial and spectral dimensions, enhancing the model's generative capabilities.

\subsection{Text-Grounded Class Conditioning (Textron)}

\begin{figure*}[t]
    \centering
    \includegraphics[width=0.86\linewidth]{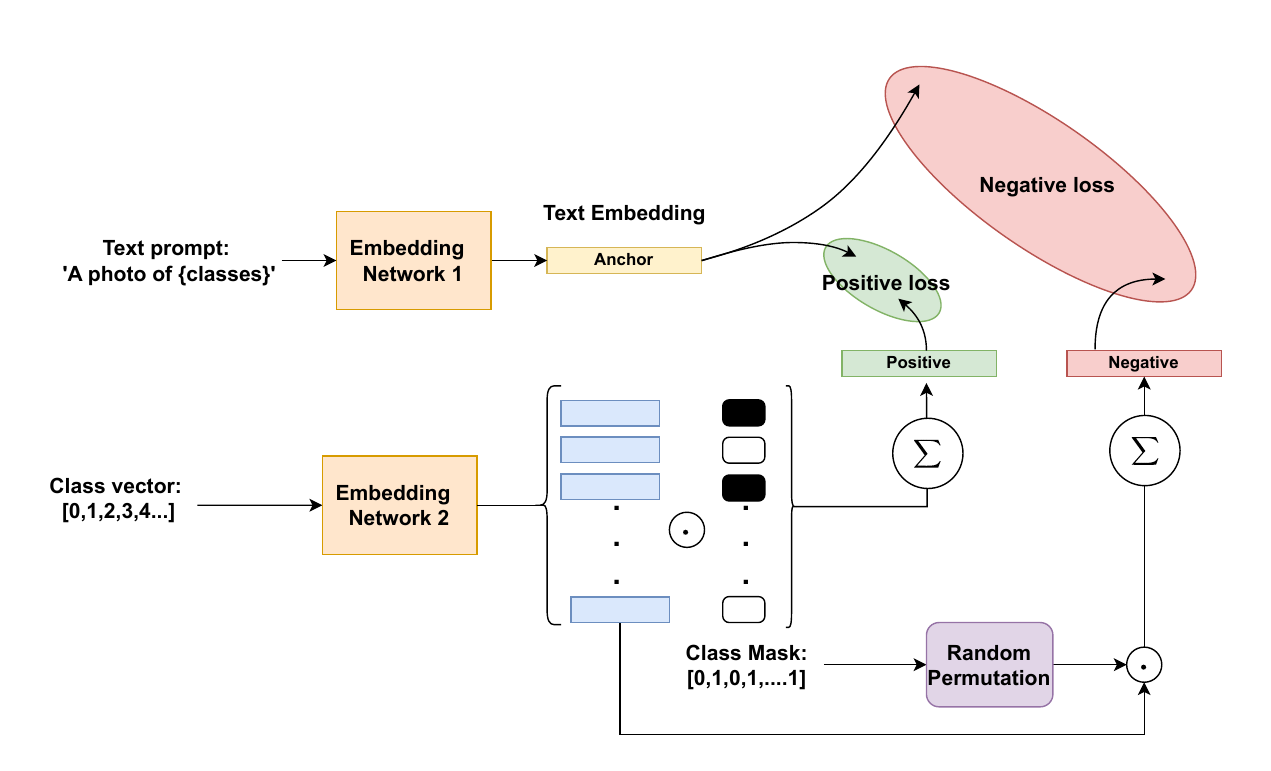}
    \caption{Textron: text-grounding of semantic class via contrastive triplet loss for text/class conditioning, enabling hot-swapping of text and class conditional input prompts during inference.}
    \label{fig:triplet}
\end{figure*}

While class embeddings $\mathbf{C}_{\text{emb}}$ can generate images and masks independently, they lack the flexibility to allow inference on text inputs directly. By aligning class embeddings with their corresponding text embeddings during training, we enable the model to perform inference using text embeddings in place of class embeddings, under the assumption that they will lie close to each other in the learned embedding space. 
Traditionally, generative models achieve text conditioning by contrasting features of generated images with their respective text embeddings, learning a similarity metric. Although effective for text conditioning, this method does not facilitate “hot-swapping” the class encoder with a text encoder during inference. 
To overcome this limitation, we implement {Textron}, based on a triplet loss using Eqn[\ref{eq:ltrip}]. In this approach \cref{fig:triplet}, the text embedding $\mathbf{T}_{\text{emb}}$ serves as the anchor, with corresponding class embeddings $\mathbf{C}_{\text{emb}}$ as the positive examples, and random permutations $\tilde{\mathbf{C}}_{\text{emb}}$ as negative examples. The triplet loss optimizes the Euclidean distance metric between these embeddings, encouraging the class and text embeddings to align.
By minimizing this loss, we ensure that the model learns a shared embedding space where text and class embeddings lie close to each other, enabling smooth substitution of class embeddings with text embeddings during inference. This approach allows the model to leverage the semantic richness of text inputs, empowering text-grounded generation in an efficient, versatile manner.

\subsection{Super-Resolution Image-Mask Pair Generation}
\label{ssec:sr}

The super-resolution model, denoted by $\mathcal{SR}$, is designed to upscale a generated low-resolution (LR) image-mask pair $\mathbf{X}_{\text{LR}} \in \mathbb{R}^{6 \times 128 \times 128}$ to produce a high-resolution (HR) image-mask pair. This model employs an Efficient sub pixel CNN (ESPCNN) \cite{shi2016realtimesingleimagevideo} with an upscale factor of $2$. Let $\mathbf{x}_{\text{LR}}^{\text{gt}} \in \mathbb{R}^{3 \times h \times w}$ be the low-resolution ground truth image-mask pair, and let $\mathbf{X}_{\text{HR}}^{\text{gt}} \in \mathbb{R}^{3 \times 2h \times 2w}$ be the corresponding high-resolution ground truth pair. During training, the super-resolution process begins by adding a small Gaussian noise to $\mathbf{X}_{\text{LR}}^{\text{gt}}$, yielding $\mathbf{\tilde{X}}_{\text{LR}}^{\text{gt}}$. The model then takes this noisy input and maps it to a higher resolution as $\mathcal{SR}(\mathbf{\tilde{X}}_{\text{LR}}^{\text{gt}}) \rightarrow \mathbf{X}_{\text{HR}}^{\text{gt}} \in \mathbb{R}^{3 \times 2h \times 2w}$. The $\mathcal{SR}$ is trained across two scales, allowing it to be applied iteratively to upscale from $128 \times 128$ to $256 \times 256$ and then to $512 \times 512$.

\subsection{Loss Functions}
\subsubsection{DDPM optimization}
The loss function for the diffusion model consists of multiple components. First, we train the discriminator $\mathcal{D}$ to distinguish between the real low-resolution ground truth image-mask pair $\mathbf{x}_{\text{lr}}^{\text{gt}}$ and the denoised image-mask pair $\hat{\mathbf{x}}_t$. The objective for the discriminator is to maximize the expectation of $\mathcal{D}(\mathbf{x}_{\text{lr}}^{\text{gt}}) = 1$ and $\mathcal{D}(\hat{\mathbf{x}}_t) = 0$. Once trained, the discriminator is frozen.
The diffusion model's objective is to maximize the expectation of the clean image mask pair $\mathbf{x}_0$ given the noisy image mask pair $\mathbf{x}_t$, the text embedding $\mathbf{T}_{\text{emb}}$, the condition class embedding $\mathbf{C}_{\text{emb}}$, and the timestep $t$. The standard diffusion loss is:
\begin{equation} \label{eq:ldiff}
\mathcal{L}_{\text{diff}} =  \lVert \epsilon - \epsilon_\theta(x_t, t, \mathbf{T}_{\text{emb}}, \mathbf{C}_{\text{emb}})\rVert ^2
\end{equation}
where $\theta = [ \theta_T, \theta_{CE}, \theta_{U}]$. 
Additionally, we include a triplet loss to ensure that the positive condition class embedding $\mathbf{C}_{\text{emb}}$ is brought closer to the anchor timestep embedding $\mathbf{T}_{\text{emb}}$, while a random permutation $\tilde{\mathbf{C}}_{\text{emb}}$ of $\mathbf{C}_{\text{emb}}$ is considered negative and is pushed away from the anchor by a margin of $1$. The triplet loss can be expressed as:
\begin{equation} \label{eq:ltrip}
\mathcal{L}_{\text{trip}} = \max ( 0, \lVert \mathbf{T}_{\text{emb}} - \mathbf{C}_{\text{emb}}\rVert^2 - \lVert \mathbf{T}_{\text{emb}} - \tilde{\mathbf{C}}_{\text{emb}}\rVert^2 + 1 )
\end{equation}

\subsubsection{Adversarial learning}
A discriminator, denoted as $\mathcal{D}$, is used to regularize the generation of image-mask pairs by distinguishing between real and denoised pairs. The discriminator is trained on real image-mask pairs versus denoised ones at a sampled timestep $t$.

Let $\mathbf{\hat{X}}_t$ represent the denoised image-mask pair at timestep $t$. The discriminator $\mathcal{D}$ is a convolution-linear model, which processes input pairs of dimension $\mathbb{R}^{h \times w}$ and produces an output $\mathcal{D}(\mathbf{\hat{x}}_t) \in \mathbb{R}^{1 \times 1}$, where the output range is constrained to $(0, 1)$ such that $
\mathcal{D} : \mathbb{R}^{h \times w} \rightarrow (0,1)
$. The training objective for $\mathcal{D}$ is to maximize the likelihood of correctly classifying real image-mask pairs as $1$ and denoised pairs as $0$.
The adversarial loss encourages the denoised image-mask pair $\hat{\mathbf{x}}_t$ to be classified as real by the discriminator:
\begin{equation} \label{eq:ladv}
\mathcal{L}_{\text{adv}} = 1 - \mathcal{D}(\hat{\mathbf{x}}_t)
\end{equation}

\noindent\textbf{Combined loss}: The total loss for the diffusion model is the sum of the diffusion loss, the triplet loss, and the adversarial loss. 

We multiply the adversarial loss by a regularization factor $\beta = 0.1$ to control its influence:
\begin{equation} \label{eq:tot}
\mathcal{L}_{\text{total}} =  \mathcal{L}_{\text{diff}} + \mathcal{L}_{\text{triplet}} + \beta \cdot \mathcal{L}_{\text{adv}}
\end{equation}
Thus the combined loss in \cref{eq:tot} optimizes the parameters in $\theta$ for conditional image generation using \cref{eq:ldiff}, text and class alignment using \cref{eq:ltrip}. The loss is also regularised using a regularized factor of \cref{eq:ladv}.

\subsubsection{Super resolution optimization}

The super-resolution model $\mathcal{SR}$ is optimized by minimizing a combination of Mean Squared Error (MSE) loss and perceptual loss on the predicted and ground truth image-mask pairs. Let $\mathbf{X}_{\text{HR}}^{\text{gt}}$ denote the high-resolution ground truth image-mask pair, and $\hat{\mathbf{X}}_{\text{HR}}$ denote the high-resolution prediction from the super-resolution model. 
\begin{equation}\label{eq: mse}
\mathcal{L}_{\text{MSE}} = \frac{1}{N} \sum_{i=1}^{N} \left( \mathbf{X}_{\text{HR}}^{\text{gt}} - \hat{\mathbf{X}}_{\text{HR}} \right)^2
\end{equation}
In addition to mean squared error (MSE) from \cref{eq: mse}, we incorporate a perceptual loss to capture high-level visual features, improving the quality of texture and structure in the generated high-resolution output. We compute the perceptual loss $\mathcal{L}_{\text{perc}}$ by measuring the similarity between the VGG-encoded feature maps of the ground truth and predicted images at level 6 using \cref{eq: perc}:
\begin{equation}\label{eq: perc}
\mathcal{L}_{\text{perc}} = \lVert \text{VGG}_6(\mathbf{X}_{\text{HR}}^{\text{gt}}) - \text{VGG}_6(\hat{\mathbf{X}}_{\text{HR}}) \rVert^2
\end{equation}
The total loss in \cref{eq: totalsr} is used for optimizing the super-resolution model $\mathcal{S}$ is the weighted sum of MSE and perceptual losses:
\begin{equation}\label{eq: totalsr}
\mathcal{L}_{\text{total}} = \mathcal{L}_{\text{MSE}} + \lambda \cdot \mathcal{L}_{\text{perc}}
\end{equation}
where $\lambda$ is a weighting factor to balance the contribution of the perceptual loss.

\section{Experiments}
\label{sec:experiments}

\noindent\textbf{Datasets:}
We evaluate our model on a diverse set of segmentation datasets, including PASCAL VOC~\cite{pascal-voc-2012}, MBRSC~\cite{humansintheloop2024}, BTCV~\cite{gibson_2018_1169361}, and CholecSeg8k~\cite{hong2020cholecseg8k}. These datasets were selected for their coverage of major domains and modalities, spanning camera images, remote sensing, radiology, and surgical imagery. To enhance class separability in segmentation masks, we employ a canonical form of the golden angle (the golden ratio of the Fibonacci series) in 3D RGB space. This transformation, known as the Fibonacci RGB space (F-RGB), provides a uniform distribution of segmentation classes in the RGB color space, making classes more distinguishable. All segmentation masks in the datasets are thus converted to F-RGB, where each class is mapped to its corresponding value in this space.\smallskip

\noindent\textbf{Implementation details:}
We train CoSimGen on inputs of size $128 \times 128$. The feature dimension of the residual U-Net $\mathbf{U}$ is set to 64, with feature multipliers of $1, 2, 4,$ and $8$. The super-resolution model $\mathcal{SR}$ is trained at resolutions of $256 \times 256$ and $512 \times 512$. For optimization, we use the Adam optimizer with a learning rate of $2 \times 10^{-4}$ and a batch size of 24. The models are implemented in PyTorch, leveraging mixed precision training to optimize computational efficiency. Training is conducted on NVIDIA H100 GPUs to ensure high-performance processing.\smallskip

\noindent\textbf{Baselines:} Given the novel challenge of entangled generation of image-mask pairs, we select both regression-based and adversarial algorithms as baselines. For adversarial approaches, we adapt classical algorithms, specifically TGAN~\cite{reed2016generative} and Pix2PixGAN~\cite{isola2017image}. As a regression-based baseline, we employ a modified version of a conditional convolutional Variational Autoencoder (VAE)~\cite{kingma2013auto}. All hyperparameters for these models are optimized to align with the explored datasets. The chosen baselines aim to highlight differences in stability and convergence between adversarial and regression-based models. This comparison is model-agnostic and serves to illustrate CoSimGen’s unique approach as a regression-based model, where it directly regresses to the noise present in the input data.\smallskip

\noindent\textbf{Evaluation protocols:}
To ensure a fair comparison, we sample images for the baselines and our model at their optimal convergence points, capturing the best performance of each model. For realistic image quality assessment, we evaluate with Fréchet Inception Distance (FID~\cite{yu2021frechet}), Kernel Inception Distance (KID~\cite{yu2021frechet}), and Inception VGG Distance~\cite{yu2021frechet}, computed between real and generated images. For alignment of generated masks with images and correctness of generated entities within an image, we compute Semantic FID (sFID~\cite{bhat2025simgen}) of semantic image regions averaging them over multiple samples. 
sFID metric refines the traditional FID by introducing a semantic perspective: it computes FID scores for individual semantic regions within the image, guided by the corresponding segmentation mask. Specifically, this involves cropping image patches based on each mask class region and calculating FID scores per class. This class-wise evaluation offers a detailed and elegant assessment of generation quality across distinct regions of interest.
The novelty of the sFID metric, introduced in \cite{bhat2025simgen}, lies in its ability to \textit{automatically} assess whether the generated images align with the input prompts: the metric considers and inherently compares the object classes specified in the input prompts against the semantic classes present in generated masks, which segment the corresponding generated images.
To assess the strength of conditioning, we calculate the Positive Predicted Value (PPV) for generated masks, verifying that the queried class is present in the generated masks.
For the super-resolution model $\mathcal{SR}$, we evaluate its performance by training on generated low-resolution (LR) and high-resolution (HR) image-mask pairs and testing on low-resolution inputs. 
We further conduct ablation studies on key aspects: feature dimension of $\mathbf{U}$, the role of triplet loss in semantic grounding, the impact of the discriminator $\mathcal{D}$ as a regularizer, and the contribution of spatio-spectral feature mixing.

\section{Results}
\label{sec:results}
This section presents the results of the CoSimGen framework, highlighting its performance across multiple datasets with a focus on image-mask pair generation and fidelity. 

\subsection{Image Quality}
\begin{table*}[!t]
\centering
\caption{Evaluation of the fidelity of the generated images across 3 datasets in comparison with the baselines across four metrics: FID, KID, VGG distance, and LPIPS distance.}
\label{tab:comparison}
\setlength{\tabcolsep}{3pt}    
\resizebox{0.98\linewidth}{!}{%
\begin{tabular}{@{}lrccccrccccrcccc@{}}
    \toprule
    \multirow{2}{*}{Model} &\phantom{abc}& \multicolumn{4}{c}{Pascal VOC} &\phantom{abc}& \multicolumn{4}{c}{MBRSC} &\phantom{abc}& \multicolumn{4}{c}{BTCV} \\
    \cmidrule{3-6} \cmidrule{8-11} \cmidrule{13-16}
    && FID & KID & VGG-D & LPIPS-D && FID & KID & VGG-D & LPIPS-D && FID & KID & VGG-D & LPIPS-D \\
    \midrule
    TGAN && 348.19 & 0.29 & 221.53 & 0.77 && 394.86 & 0.27 & 113.09 & 0.72&& 394.05& 0.53& 146.31& 0.60  \\
    Pix2PixGAN && 348.05 & 0.30 & 225.56 & 0.79 && 410.18 & 0.34 & 117.94 & 0.70 && 284.60&0.36 & 152.80& 0.54  \\
    CVAE && 337.41 & 0.35 & \textbf{204.97} & 0.76 && 326.16&0.27 &\textbf{106.79} &0.70 && 192.21&0.19 & 144.84&0.45  \\
    CoSimGen (Ours) && \textbf{206.29}& \textbf{0.20} & 227.64 & \textbf{0.74} && \textbf{203.67} & \textbf{0.11} & 110.43 & \textbf{0.63} && \textbf{159.92}&\textbf{0.13} &\textbf{139.35} & \textbf{0.53} \\
    \bottomrule
\end{tabular}
}
\end{table*}

\begin{figure}[t]
    \centering
    \includegraphics[width=1.0005\linewidth]{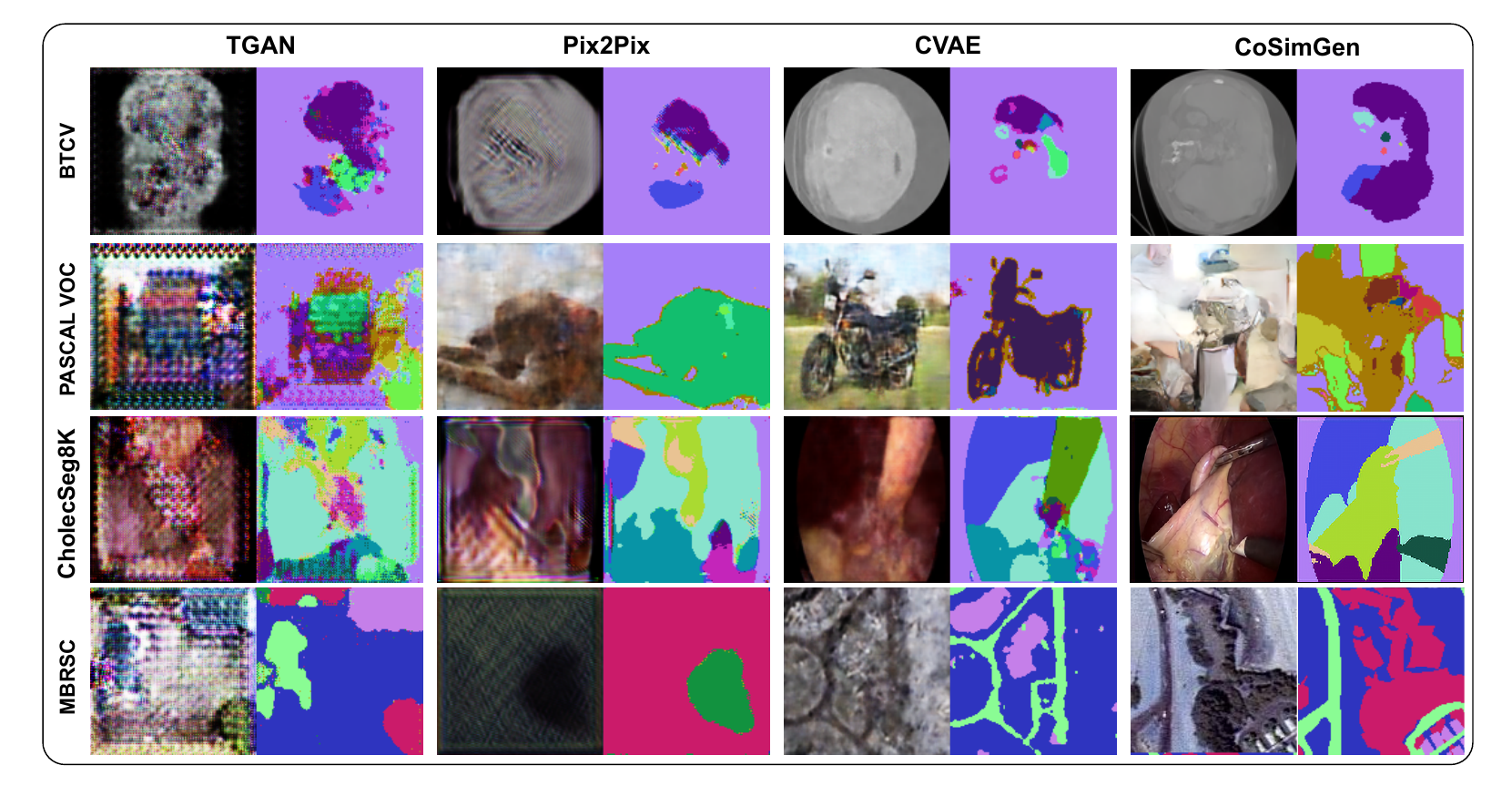}
    \caption{Qualitative comparisons of generated image-mask pairs}
    \label{fig:ourgen}
\end{figure}

Fidelity scores on the quality of the generated images in comparison with the baseline is presented in \cref{tab:comparison}. Our proposed model obtained the best FID, KID and LPIPS distances in Pascal VOC, MBRSC and BTCV datasets. Our model also obtained the best VGG distance on the BTCV dataset while The CVAE baseline was better on Pascal VOC and MBRSC datasets.
Qualitatively, \cref{fig:ourgen} provide a visual comparison of CoSimGen's outputs with baselines like CVAE, TGAN, and Pix2PixGAN. In terms of image quality, CoSimGen generates crisper, more detailed images with better mask alignment than the baselines, particularly for datasets with abundant training data.  
Performance variations are observed to be dataset-dependent, with CVAE outperforming CoSimGen on smaller datasets like PASCAL VOC due to its lower data dependency. However, CoSimGen excels with larger datasets such as CholecSeg8k, MBCV, etc., highlighting its scalability.

\cref{fig:ourgen} illustrates that the convergence of CVAE and our model, CoSimGen, is superior to that of adversarially trained models, such as TGAN and Pix2PixGAN. It is observed in \cref{fig:ourgen} that our generated images exhibit crisper details compared to those from CVAE. On radiology images, while CVAE performs comparably to our model, CoSimGen demonstrates higher fidelity in the generated masks. Overall, our model consistently outperforms other models across all datasets, achieving superior visual quality and mask alignment.
CVAE performing better than CoSimGen in PASCAL VOC in qualitative results showcases the limitation of CoSimGen as a data hungry model.
More results are provided in the Appendix.

\subsection{Image-Mask Alignment}
% Here we talk about the table result of image and mask fidelity using metrics such as semantic FID and PPV. 

\begin{table}[!t]
\centering
\caption{Evaluation of the generated mask-image alignment in 3 datasets in comparison with the baselines across two metrics: semantic fréchet inception distance (sFID) and positive predicted value (PPV).}
\label{tab:class_fid_ppr_comparison}
\setlength{\tabcolsep}{3pt}    
\resizebox{0.98\linewidth}{!}{%
\begin{tabular}{@{}lrccrccrcc@{}}
    \toprule
    \multirow{2}{*}{Model} &\phantom{abc}& \multicolumn{2}{c}{Pascal VOC} &\phantom{abc}& \multicolumn{2}{c}{MBRSC} &\phantom{abc}& \multicolumn{2}{c}{BTCV} \\
    \cmidrule{3-4} \cmidrule{6-7} \cmidrule{9-10}
    && sFID & PPV && sFID & PPV && sFID & PPV\\
    \midrule
    TGAN && 348.61& \textbf{1.0} && 431.12& 0.43&& 406.07& 0.50 \\
    Pix2PixGAN && \textbf{326.66} & 0.81 && 462.74& 0.84&& 323.26& 0.43 \\
    CVAE && 381.83 & 0.91 && 422.80&\textbf{0.91} && 250.52&\textbf{0.59}\\
    CoSimGen (Ours) && 343.66 & 0.78 && \textbf{294.68} & 0.87 && \textbf{198.74} & 0.35 \\
    \bottomrule
\end{tabular}
}
\end{table}

\cref{tab:class_fid_ppr_comparison} shows the results on the assessment of the fidelity of image-mask alignment. SimGen show the best alignment in terms of sFID on MBRSC and BTCV datasets, second best in terms of PPV on the MBRSC and BTCV datasets. However, it was worse on Pascal VOC which is connected to the small size of the dataset. 
Our model outperforms the baselines on most datasets and most metrics \cref{tab:comparison} and \cref{tab:class_fid_ppr_comparison}. The PPV of TGAN is high as it is generating similar images most of the times. CVAE is better on PASCAL VOC suggesting vaes are better with lower data however our model outperforms when the datasets are larger.

\subsection{Input-Output Alignment}

\begin{figure}[t]
    \centering
    \includegraphics[width=1.0\linewidth]{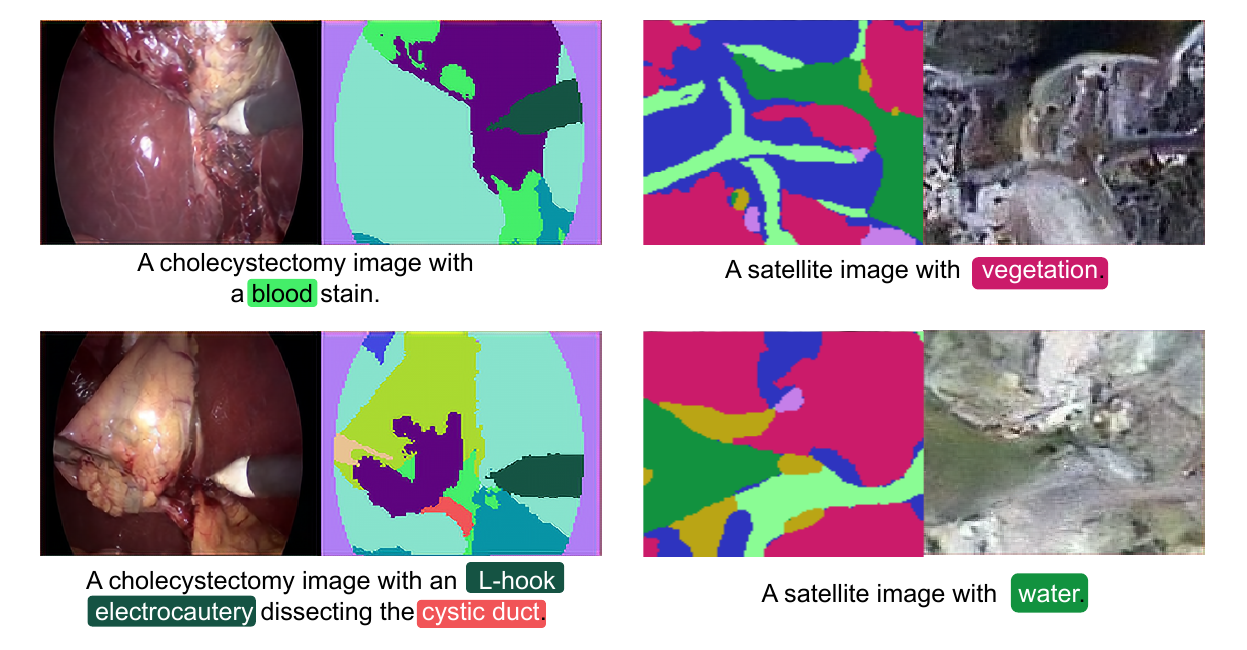}
    \caption{Qualitative results showing text(class)-conditioned image-mask generation}
    \label{fig:conditioned-gen}
\end{figure}

% {\color{red}
The fidelity of the generated image-mask pairs with respect to the input prompts is assessed also using sFID in \cref{tab:class_fid_ppr_comparison}. In this case, the metric scores the model lower when classes specified in the input prompts are missing in the generated mask classes. Qualitative results in \ref{fig:conditioned-gen} shows the alignments between input prompts and generated output pairs.
More results are provided in the Appendix.
% }

\subsection{High-Resolution (HR) Outputs}

\begin{figure}[t]
    \centering
   \includegraphics[width=.99\linewidth]{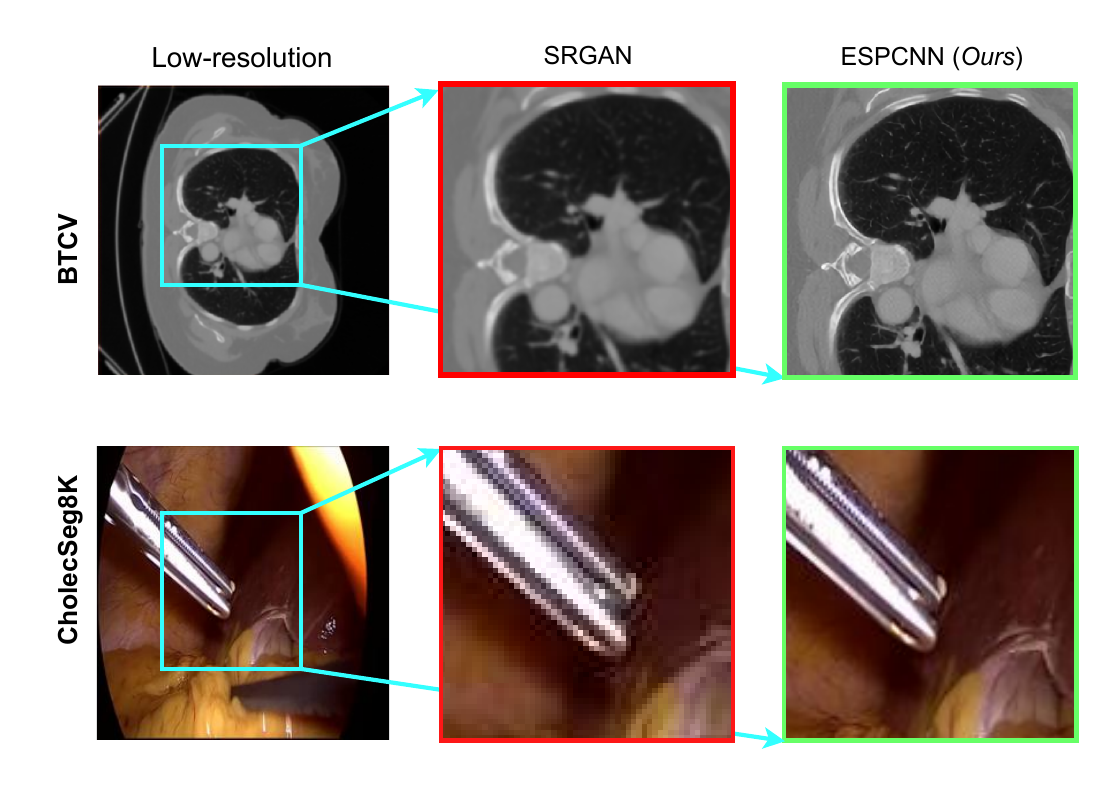}
    \caption{Qualitative comparison of super-resolution result of ESPCNN (in our model) and SRGAN (baseline).}
    \label{fig:lrhr}
\end{figure}

The super-resolution (SR) images produced by ESPCNN~\cite{shi2016realtimesingleimagevideo}, utilized in our proposed CoSimGen framework, are compared with baseline outputs from SRGAN~\cite{ledig2017photo} on CholecSeg8K~\cite{hong2020cholecseg8k} and BTCV~\cite{gibson_2018_1169361} datasets. 
The results in \cref{fig:lrhr} demonstrate that ESPCNN effectively captures high-frequency details that SRGAN fails to reproduce. This distinction is particularly evident in the sharper boundaries between textures, such as those of organs, bones, and blood vessels, highlighting ESPCNN's superior ability to preserve structural details.
More results are provided in the Appendix.

\subsection{Ablation Study}

\begin{figure}[!t]
    \centering
    \includegraphics[width=0.998\linewidth]{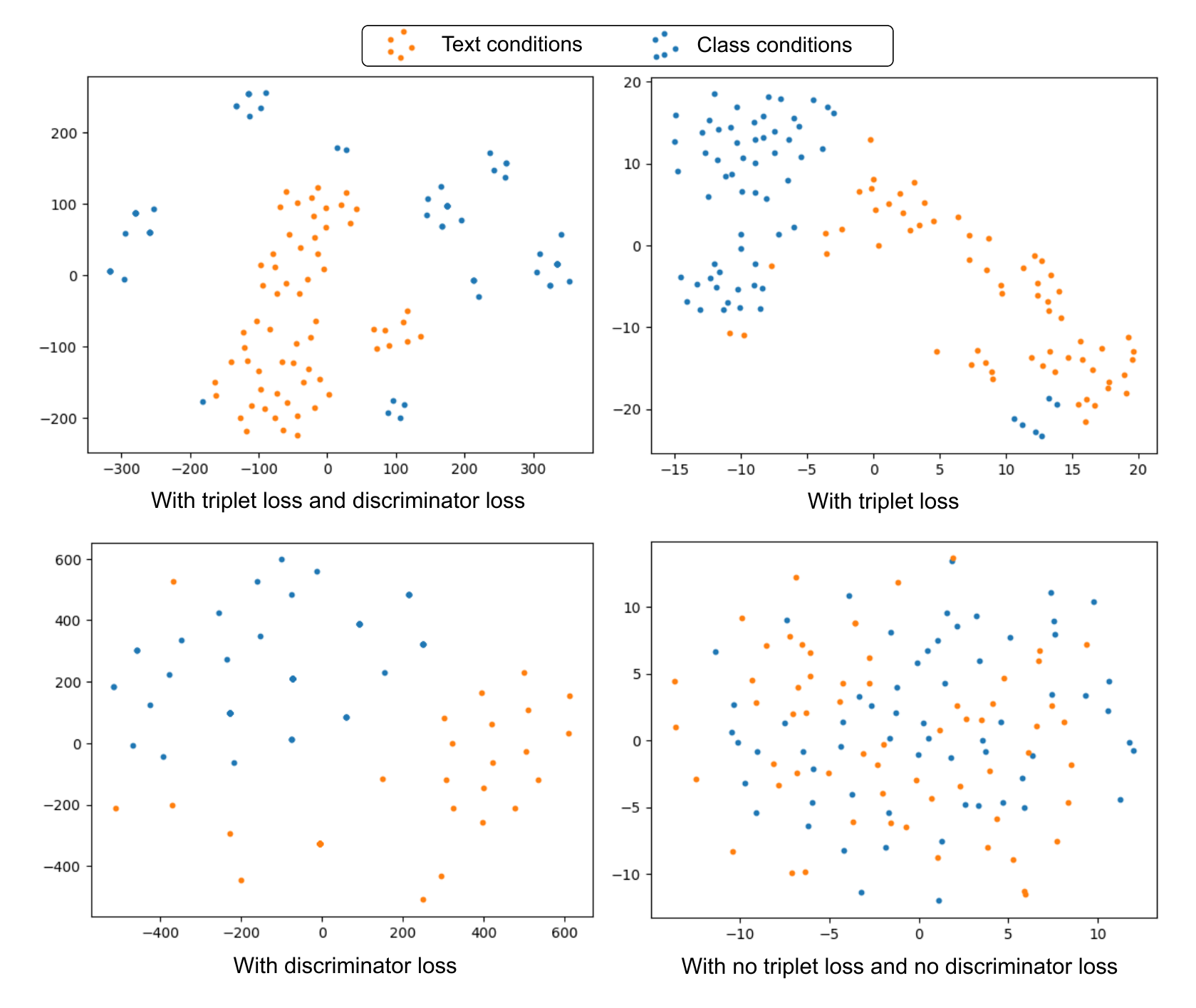}
    \caption{Ablation on the contributions of using disriminator loss and triplet loss on MBRSC dataset.}
    \label{fig:ablation-loss}
\end{figure}

CoSimGen introduces two key innovations compared to traditional approaches: (1) the use of triplet loss for spatio-spectral conditioning grounded on text embeddings, and (2) the inclusion of discriminator loss as a regularizer. We conduct an ablation study to assess the impact of these contributions individually and in combination. 
Ablation results in \cref{fig:ablation-loss} examines the impact of triplet loss and show that it enhances text-grounded conditioning, improving class FID.  
It also reveals that discriminator loss improves fidelity by regularizing the output distribution. 
We also observed that the combining both losses achieves the highest performance, showcasing the complementary benefits of both losses in text alignment and image fidelity.

Estimating the distribution of paired image and mask pairs is a challenging task using an adversarial learning method, as it requires modeling a joint continuous and discrete distribution. This complexity contributes to high instability in adversarial models for this task. In contrast, a regression-based model minimizes the error between the ground truth and the prediction using residual losses, resulting in a smoother optimization landscape. 
More ablation studies are provided in the Appendix.

\subsection{Discussion}

Our experiments reveal that models optimized with regression objectives exhibit greater stability compared to those with adversarial objectives, particularly in tasks involving joint estimation of continuous and discrete distribution pairs. In practice, we observed that adversarial models frequently suffered from mode collapse, which undermines their reliability for such tasks. By contrast, regression-based approaches minimize prediction error in a smoother optimization landscape, which contributes to improved stability. We also found that incorporating adversarial loss as a regularizer in our model introduced oscillatory behavior in early training stages, where generation quality fluctuates. However, as training progresses, these oscillations diminishes, and generation quality stabilizes. This suggests that adversarial loss, while beneficial as a regularizer, may require careful tuning to balance stability with generation fidelity.
\smallskip

\noindent\textbf{Limitations:} Diffusion models are inherently data-hungry, and acquiring annotated segmentation masks paired with class-specific text labels is costly. Our experiments show that a reduced dataset size impairs generation quality. When dataset size was artificially increased using suitable augmentations (e.g., random rotations and flips, which work effectively for geospatial datasets but are less applicable to everyday objects), we observed a significant boost in model performance.
This data dependency poses a major limitation for the development of diffusion models at a large scale, especially in domains with limited annotated data. Thus, exploring generative models in few-shot settings becomes a promising direction for future work.

\section{Conclusion}  \label{sec:conclusion}
This work introduces {CoSimGen}, a novel diffusion-based framework for controllable simultaneous image and segmentation mask generation. By addressing the critical challenges in existing generative models, {CoSimGen} provides a unified solution for producing high-quality paired datasets with precise control during generation. The model leverages text-grounded class conditioning, spatial-temporal embedding fusion, and multi-loss optimization, enabling robust performance across applications requiring spatial accuracy and flexibility.  
{CoSimGen} demonstrates state-of-the-art performance on diverse datasets, making it a versatile tool for augmenting datasets, simulating rare scenarios, and tackling domain-specific challenges. Its outputs offer a scalable alternative to manual annotation, significantly reducing the time and resources required for dataset creation. Moreover, the generated paired data serve as a ready source for pretraining models, given the framework’s ability to produce an unlimited variety of high-fidelity, condition-adherent examples.  
Beyond its utility in dataset augmentation, {CoSimGen} establishes a foundation for future research in multi-modal, multi-class, and domain-adaptive generative frameworks. By bridging the gap between generative AI and real-world applications, the framework addresses critical bottlenecks in precision-driven and privacy-sensitive domains, advancing cross-domain AI research and deployment.  
{CoSimGen} represents a significant step forward in enabling scalable, controllable data generation, unlocking new possibilities for pretraining, robustness testing, and real-world impact.

\subsection*{Acknowledgements}
This work was supported by French state funds managed within the Plan Investissements d’Avenir by the ANR under grants: ANR-20-CHIA-0029-01 (National AI Chair AI4ORSafety), ANR-22-FAI1-0001 (project DAIOR), ANR-10-IAHU-02 (IHU Strasbourg). This work was granted access to the servers/HPC resources managed by CAMMA, IHU Strasbourg, Unistra Mesocentre, and GENCI-IDRIS [Grant 2021-AD011011638R4].

%%%%%%%%%%%%%%%%%%%% REFERENCES %%%%%%%%%%%%%%%%%%%%%%

% \bibliographystyle{unsrtnat}
% \bibliographystyle{splncs04}
\bibliographystyle{IEEEtran}
\bibliography{arxiv}

\onecolumn
\clearpage
\setcounter{page}{1}
% \maketitlesupplementary
\appendix
\section*{\textsc{Appendix}}
\section{Further Results and Discussions}
\begin{figure*}[!ht]
    \centering
    \includegraphics[width=0.98\linewidth]{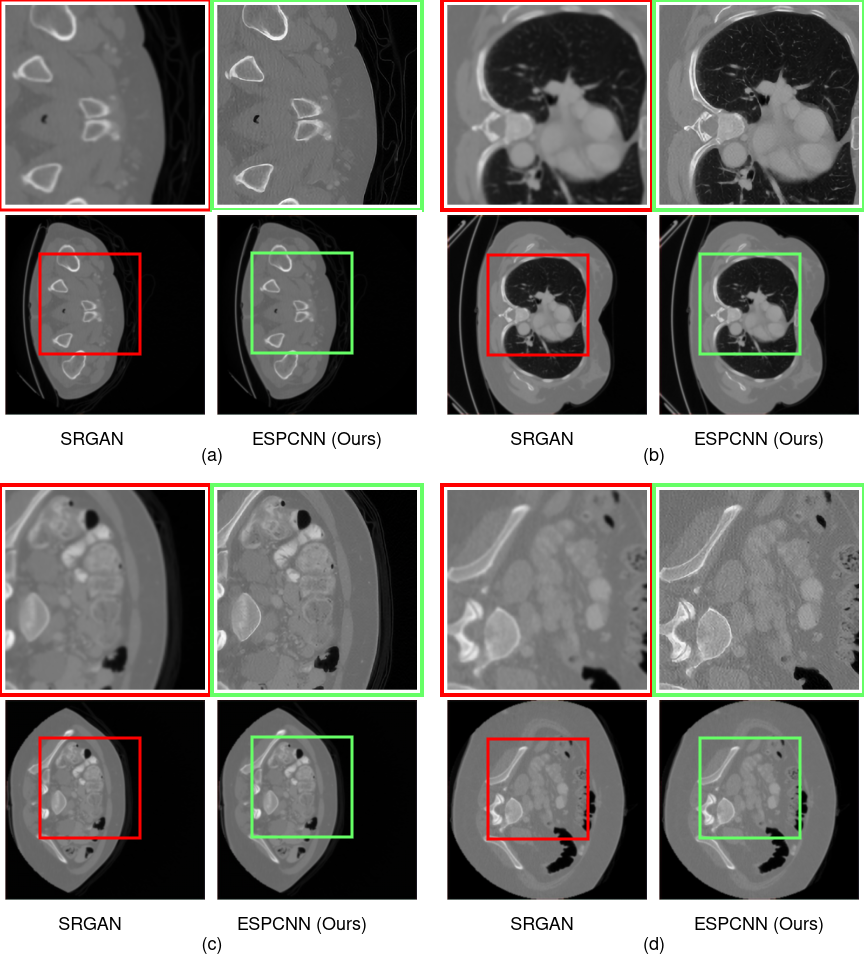}
    \caption{Qualitative results on BTCV dataset highlighting the visualization of different textures of in the data (a) nerves, muscles and bones, (b) bronchi, heart and aorta, (c) colon, muscles, bones and gut, and (d) colon, gut and bones.
    For each image, we show the generated low resolution image (bottom) along with the associated super-resolution image of SRGAN (top-left), and super-resolution image of ESPCNN (top-right). We also observe that the low resolution images have artifacts as they are generated, however as the high resolution images have good fidelity, thus it does not impact the interpretation of the CT images.}
    \label{fig:btcvsup}
\end{figure*}

In addition to the low-resolution (LR) qualitative images presented in the main text, we provide more low- and high-resolution outputs generated by our proposed model. \cref{fig:btcvsup} showcases extensive qualitative results on the BTCV dataset~\cite{gibson_2018_1169361}. The super-resolution (SR) images produced by ESPCNN~\cite{shi2016realtimesingleimagevideo}, utilized in our proposed CoSimGen framework, are compared with baseline outputs from SRGAN~\cite{ledig2017photo}. 
The results demonstrate that ESPCNN effectively captures high-frequency details that SRGAN fails to reproduce. This distinction is particularly evident in the sharper boundaries between textures, such as those of organs, bones, and blood vessels, highlighting ESPCNN's superior ability to preserve structural details.

\begin{figure*}[!ht]
    \centering
    \includegraphics[width=0.98\linewidth]{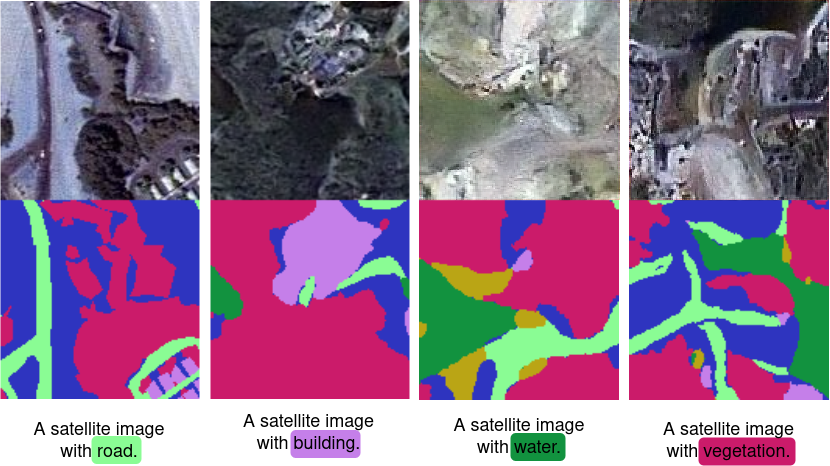}
    \caption{Additional qualitative results on the MBRSC dataset showcasing CoSimGen's image-mask outputs generated based on text prompts describing various semantic classes. The displayed images are cropped to a resolution of $128\times128$ owing to the extensive size of satellite imagery. We observe that the class corresponding to the queried prompt is always present in the generated semantic mask.}    
    \label{fig:mbrscsup}
\end{figure*}

In \cref{fig:mbrscsup}, we present additional qualitative results of CoSimGen on the MBRSC dataset~\cite{humansintheloop2024}, highlighting its ability to generate satellite images that accurately reflect the semantic classes queried in the text prompts. These results demonstrate CoSimGen's capability to conditionally align the generated content with specified semantic details, effectively capturing and representing features such as buildings, roads, water bodies, vegetation, and other land use patterns. This level of precision is particularly significant for applications requiring detailed spatial representations, such as urban planning or environmental monitoring. 

\begin{figure*}[!ht]
    \centering
    \includegraphics[width=0.98\linewidth]{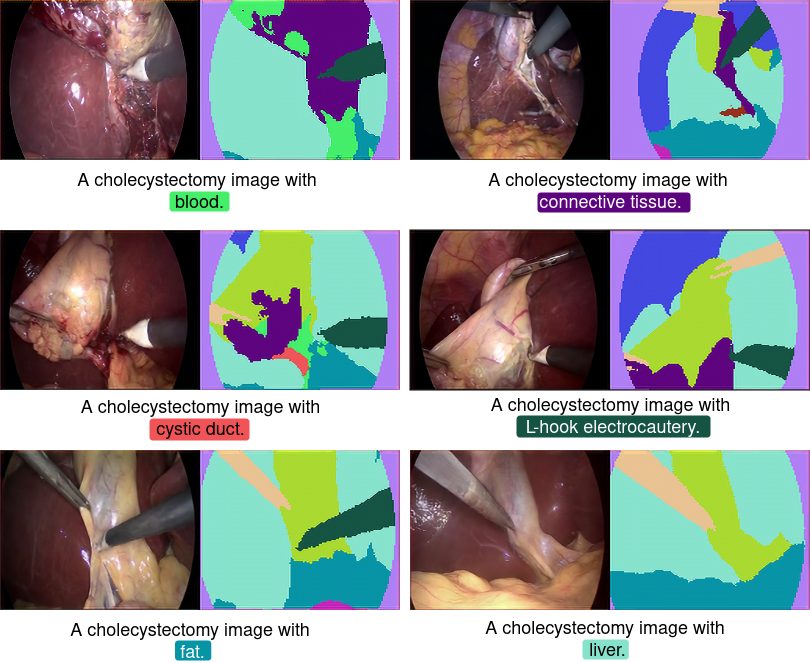}
    \caption{Qualitative results of CoSimGen on Cholecseg8k dataset. We observe that major portions of the image is covered by liver/ gallbladder along with the prompted class. Further we observe that the queried class is semantically present in the generated mask and image.}
    \label{fig:cholecseg8kcsup}
\end{figure*}

\cref{fig:cholecseg8kcsup} showcases qualitative outputs of CoSimGen on the Cholecseg8k dataset, including SR images and their corresponding segmentation masks. The SR images ($512\times512)$ are resized to actual resolution ($480\times854$) of images in the dataset for improved visualization. CoSimGen's generated mask captures the presence of the semantic class queried in the input prompt.

\begin{figure*}[!ht]
    \centering
    \includegraphics[width=0.98\linewidth]{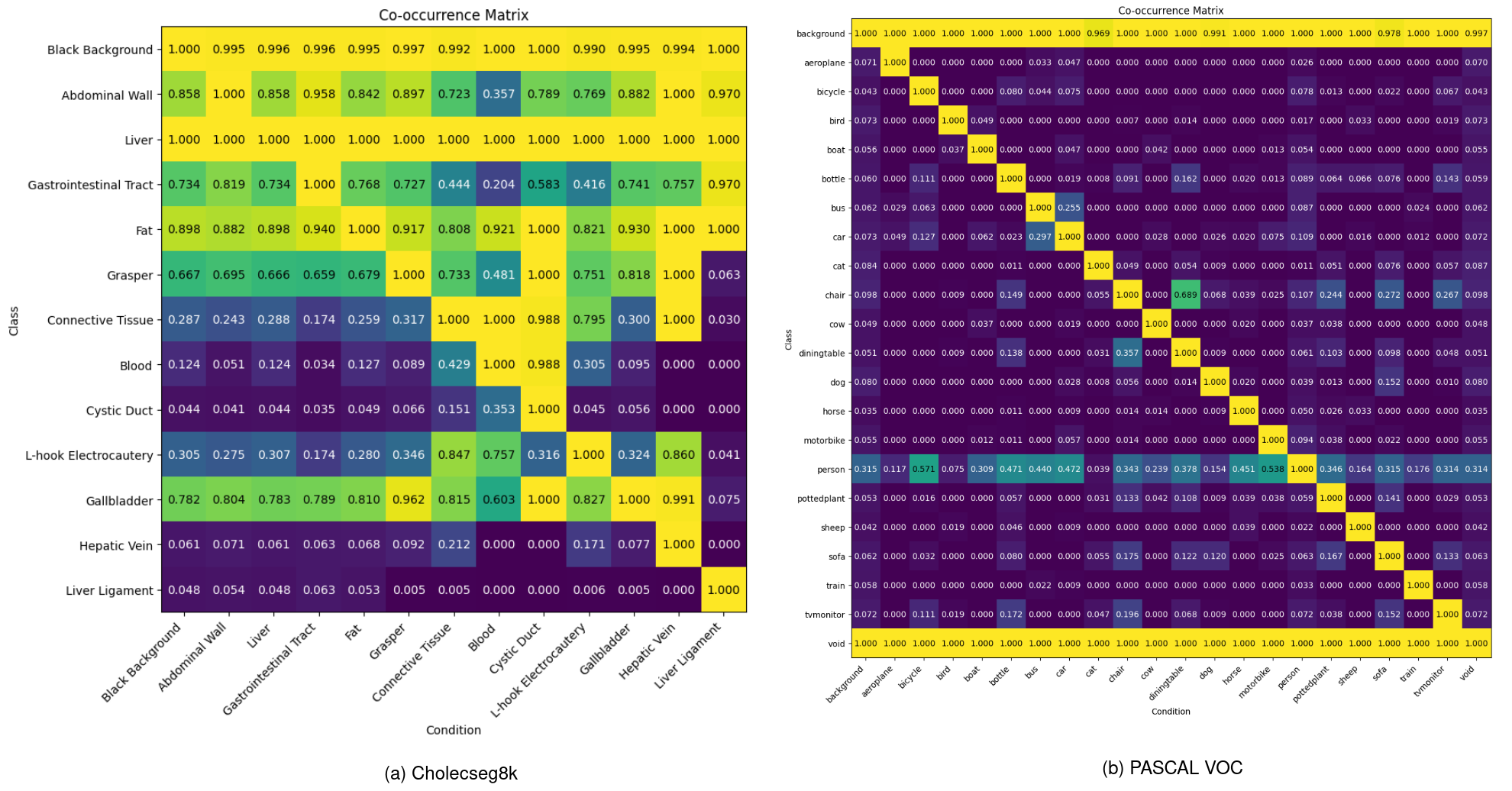}
    \caption{Class condition co-occurrence matrix for Cholecseg8k and PASCAL VOC. We observe that Cholecseg8k has a very dense co-occurence matrix whereas the PASCAL VOC has a sparse matrix. Classes such as gallbladder and liver have a larger overall fooprint in the image.}
    \label{fig:co-occurencepascal}
\end{figure*}

\begin{figure*}[!ht]
    \centering
    \includegraphics[width=0.98\linewidth]{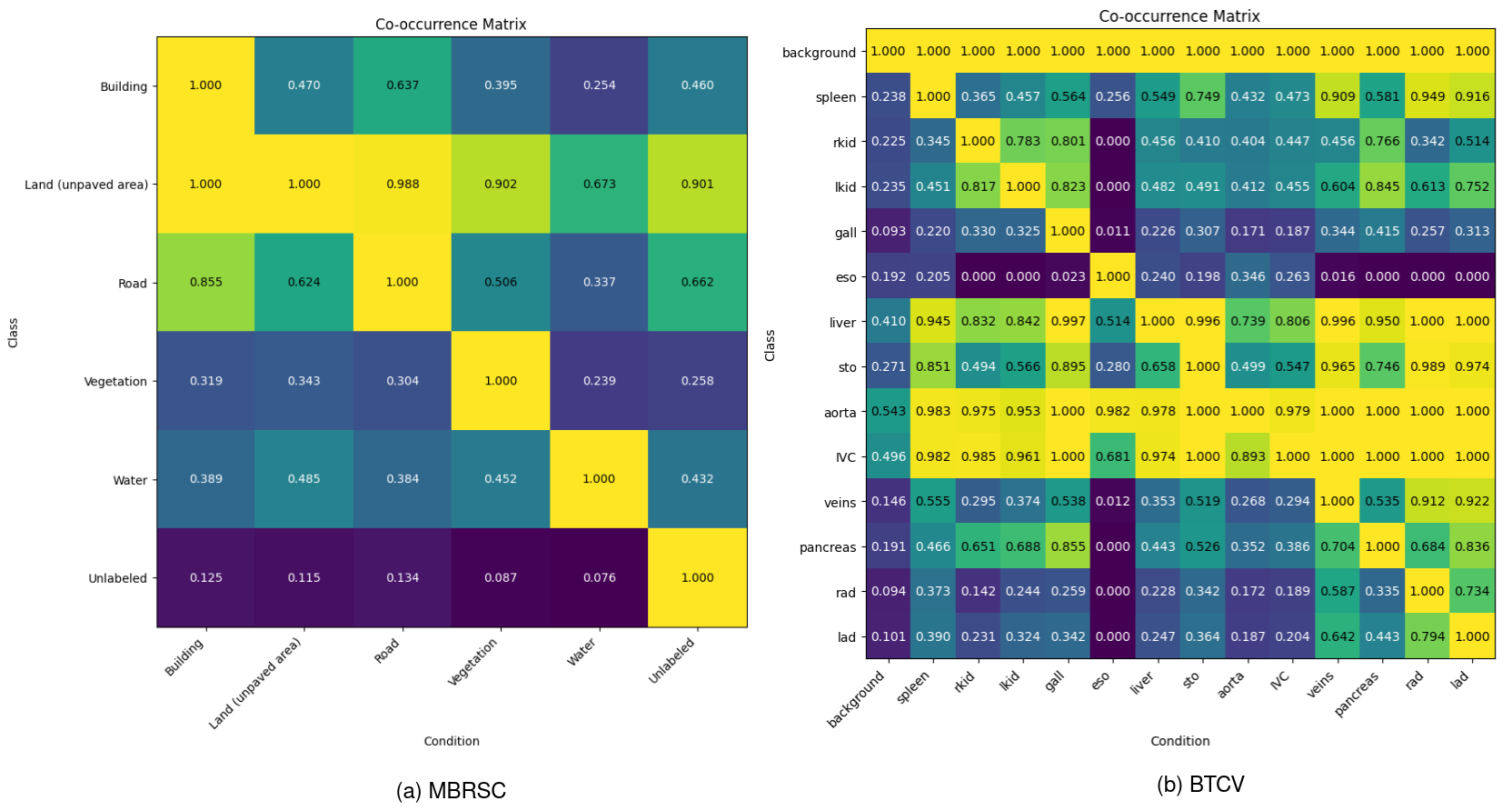}
    \caption{Class condition co-occurrence matrix for MBRSC and BTCV. We observe that they have a more balanced co-occurrence matrix. BTCV matrix differs from Cholecseg8k in one key factor. BTCV being slices of 3D volume, classes like aorta and IVC (inferior vena cava are always co-occuring but have a lower footprint on the image overall. }
    \label{fig:co-occurencebtcv}
\end{figure*}

Class condition co-occurrence matrix in \cref{fig:co-occurencepascal} and \cref{fig:co-occurencebtcv} describes the probability of a class occurring given a conditional class. A high value in the matrix indicates that the co existence of a class and a condition, which is accepted as the inherent outcome. \cref{fig:co-occurencepascal} (a) shows that there is a high co-occurrence among classes and \cref{fig:cholecseg8kcsup} shows that large sections of the image are covered by liver, gallbladder and fat. Thus visual confirmation of fidelity is justified over computing computing semantic FID as the features of generated images would largely overlap with the real images, thereby making semantic FID, FID, KID, VGG distance uninformative for Cholecseg8k. \cref{fig:co-occurencepascal} (b) shows that PASCAL VOC has a very sparse matrix, indicating an extremely low overlap in class and condition overlap. This makes it harder for the model to generalize at inference times. We compare CoSimGen with CVAE, pix2pix, TGAN as they are the broad classes of adversarial and regressive models. Additionally, in \cref{fig:ourgen} (a) it is observed that CVAE and pix2pix overfits on PASCAL VOC. \cref{fig:ourgen} also highlights CoSimGen's robustness to overfitting and its superior performance given enough data. \cref{fig:co-occurencebtcv} (a,b) shows moderate overlap among the class and condition which is optimal for convergence. CoSimGen outperforms other baselines in terms of image fidelity when presented with optimum data volume .

\begin{figure}[h]
    \centering
    \includegraphics[width=1.0\linewidth]{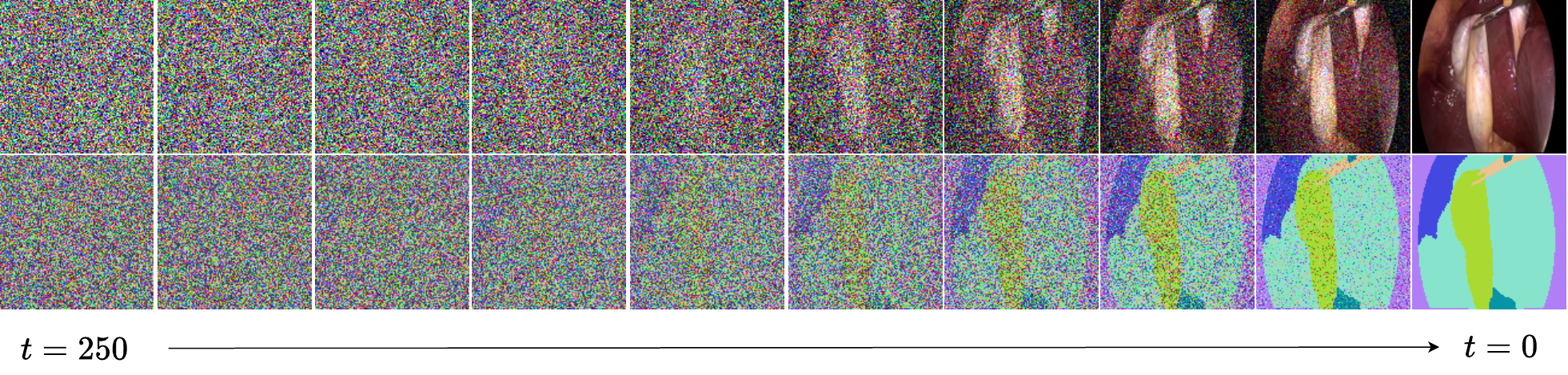}
    \caption{Predicted outputs during reverse diffusion process using CholecSeg8k dataset.}
    \label{fig:noiseschedule}
\end{figure}

% \subsection{More Discussion on the Ablation Studies}
In the main text, \cref{fig:ablation-loss} highlights the impact of incorporating text-grounded class conditioning via triplet loss and adversarial conditioning through discriminator loss in CoSimGen. The visualization shows blue dots representing class conditions and orange dots representing text conditions. When both triplet and discriminator losses are used, class conditions are observed to form distinct clusters guided by textual semantics. This indicates the model’s ability to align textual and class embeddings effectively, facilitating meaningful associations between the two.
The triplet loss encourages semantically similar class conditions to converge within the latent space based on their shared textual context. Meanwhile, the discriminator loss ensures that the generated outputs align closely with real data distributions, enhancing overall fidelity. Together, these mechanisms enable the model to generate contextually coherent image-mask pairs that remain consistent with both the textual and class conditioning.
This alignment not only showcases the synergy between the two loss functions but also provides practical flexibility. During inference, the model can effectively use either text prompts or class vectors to generate high-quality outputs. This flexibility is crucial for applications where one form of conditioning might be more readily available or practical than the other, demonstrating the robustness and adaptability of the proposed framework.\\\

To summarize, CoSimGen demonstrates the ability to generate high-fidelity image-mask pairs that align with user-provided prompts. The generated masks consistently include the semantic class specified in the prompt while also reasoning contextually to incorporate additional relevant and complementary classes, resulting in more meaningful and realistic outputs. Non-required classes for downstream usage can be easily removed via simple postprocessing, as the mask’s class identities are directly mappable to the dataset's class names.
CoSimGen addresses the challenges of image-mask pair generation across diverse datasets by leveraging its robust generative modeling framework. It achieves high accuracy using metrics such as semantic FID, FID, KID, and VGG distance, alongside visual fidelity assessments. The model's scope lies in efficiently generating image-mask pairs from large datasets, as evidenced by its performance on datasets like CT scans, surgical images, satellite images, and natural images. 
As a limitation, the model's output quality is suboptimal with small size datasets as seen in Pascal VOC (\cref{fig:ourgen}). As a bottleneck, there are limited meaningful augmentation to artificially increase the size of the dataset without creating unrealistic images for input prompts.
Furthermore, the use of a super-resolution model ( \cref{ssec:sr}) enhances the fidelity of outputs, capturing finer details and improving the quality of high-resolution images.
The paper also highlights CoSimGen's strong mathematical foundation, enabling spatio-spectral feature mixing and text-grounded conditioning. Ablation studies (\cref{fig:ablation-loss}) showcase the importance of text-guided alignment and adversarial discriminator loss in improving the model’s capability to cluster similar conditions and generate contextually relevant outputs. These findings, combined with a detailed analysis of quantitative and qualitative results, make this work an invaluable contribution to controllable image-mask pair generation.

\end{document}